\documentclass[5p,times]{elsarticle}
\usepackage{hyperref}
\usepackage{lineno}
\usepackage{multirow}
\usepackage{multicol}
\usepackage{makecell}
\usepackage{algorithm}
\usepackage{algorithmic}
\usepackage{graphicx}
\usepackage{subfigure}
\usepackage{longtable}
\usepackage{booktabs}
\usepackage{tabularx}
\biboptions{sort&compress}

\usepackage{amssymb}

\usepackage{amsmath}

\journal{Future Generation Computer Systems}

\begin{document}

\begin{frontmatter}



\title{Causal invariant geographic network representations with feature and structural distribution shifts}


\author[csu]{Yuhan Wang}
\author[csu]{Silu He}
\author[csu]{Qinyao Luo}
\author[csu]{Hongyuan Yuan}
\author[csu]{Ling Zhao}
\author[csu1]{Jiawei Zhu}
\author[csu]{Haifeng Li\corref{cor1}}
\ead{lihaifeng@csu.edu.cn}
\cortext[cor1]{Corresponding author.}

\affiliation[csu]{organization={Department of Geographic Information System, Central South University},
            addressline={}, 
            city={Changsha},
            postcode={410000}, 
            state={Hunan},
            country={China}}
\affiliation[csu1]{organization={School of Architecture and Art, Central South University}, 
            addressline={}, 
            city={Changsha},
            postcode={410083}, 
            state={Hunan},
            country={China}}

\begin{abstract}
Relationships between geographic entities, including human-land and human-people relationships, can be naturally modelled by graph structures, and geographic network representation is an important theoretical issue. The existing methods learn geographic network representations through deep graph neural networks (GNNs) based on the i.i.d. assumption. However, the spatial heterogeneity and temporal dynamics of geographic data make the out-of-distribution (OOD) generalisation problem particularly salient. We classify geographic network representations into invariant representations that always stabilise the predicted labels under distribution shifts and background representations that vary with different distributions. The latter are particularly sensitive to distribution shifts (feature and structural shifts) between testing and training data and are the main causes of the out-of-distribution generalisation (OOD) problem. Spurious correlations are present between invariant and background representations due to selection biases/environmental effects, resulting in the model extremes being more likely to learn background representations. The existing approaches focus on background representation changes that are determined by shifts in the feature distributions of nodes in the training and test data while ignoring changes in the proportional distributions of heterogeneous and homogeneous neighbour nodes, which we refer to as structural distribution shifts. We propose a feature-structure mixed invariant representation learning (FSM-IRL) model that accounts for both feature distribution shifts and structural distribution shifts. To address structural distribution shifts, we introduce a sampling method based on causal attention, encouraging the model to identify nodes possessing strong causal relationships with labels or nodes that are more similar to the target node. This approach significantly enhances the invariance of the representations between the source and target domains while reducing the dependence on background representations that arise by chance or in specific patterns. Inspired by the Hilbert–Schmidt independence criterion, we implement a reweighting strategy to maximise the orthogonality of the node representations, thereby mitigating the spurious correlations among the node representations and suppressing the learning of background representations. In addition, we construct an educational-level geographic network dataset under out-of-distribution (OOD) conditions. Our experiments demonstrate that FSM-IRL exhibits strong learning capabilities on both geographic and social network datasets in OOD scenarios.
\end{abstract}




\end{frontmatter}



\section{Introduction}
\label{section1}
Due to the diversity of data, the complexity of relationships and the wide range of applications involving geographic networks, modelling and analysis methods for geographic networks have long been important research components in geographic information science\cite{1,2,3,4}. The development and improvement of modelling and analysis methods for geographic networks can help solve real-world problems such as urban planning, traffic congestion and disease transmission\cite{5,6}.

In geographic networks, large numbers of entities are often involved, along with complex relationships between these entities that contain a wealth of attributes and interaction information. These entities include but are not limited to geographic locations, populations, or other geographically relevant factors, and their relationships span spatial connectivity, social interactions, and economic transactions, among others. This complexity often makes traditional data analysis methods inadequate. Therefore, we need to model geographic networks with graph structures because the core value of graph data lies in their ability to effectively capture and represent complex spatial relationships. In recent years, graph neural networks (GNNs) have made significant progress as a key technology in the field of geographic network-based graph data analysis\cite{7,8,9,10}. The core idea of graph neural networks is to combine the information of each node with that of its neighbours to obtain a richer and more comprehensive representation of the input data\cite{11,12,13,14}. Recently, GNNs have produced remarkable results in various geographic applications, such as traffic flow prediction\cite{15}, urban planning and management\cite{16,17,18,19}.

The current graph models are primarily established under the i.i.d. assumption, meaning that both the training and test set data are independently drawn from the same distribution. However, geospatial data are often subject to a variety of factors during collection and processing, such as temporal dynamics, spatial heterogeneity, human activities, and policy changes, which can cause data deviations, i.e., \(P_{t r}(Y \mid X) \neq P_{t e}(Y \mid X)\)\cite{20,21,22,23,24,25}. For instance, in the data collection phase, data are usually gathered based on an actual real-world situation. This process can be influenced by factors such as human behaviour and technological limitations, leading to a certain bias in the collected data. This results in a deviation from the real data distribution in actual applications, contradicting the i.i.d. assumption. As a result, we encounter performance degradation when applying a trained model to data that do not satisfy the i.i.d. assumptions\cite{oodbench}. This is due to the lack of model performance on new data outside the training data distribution, which is a deficiency known as poor out-of-distribution (OOD) generalisability\cite{26,27,28,29,30,31graphood}.

To address this problem, it is crucial to learn GNNs with out-of-distribution generalisation capabilities and to achieve relatively stable performance in the presence of a distribution shift. Distribution shifts in geographic networks can occur at both the feature and topology levels. Geographic network representations acquired by GNNs based on a message passing mechanism can be classified into invariant representations ($C$) and background representations ($S$). In the face of a distribution shift, the label relationship that always remains stable is the invariant representation, which satisfies \(P_{tr}(Y=y \mid C=c) = P_{te}(Y=y \mid C=c)
\), while the label relationship that changes is the background representation, which satisfies \(P_{tr}(Y=y \mid S=s) \neq P_{te}(Y=y \mid S=s)\). Under the influence of a specific environmental/selection bias, a particular confounding variable may be overrepresented or underrepresented in the samples, which leads to a spurious correlation between the invariant representations and the background representations. In the context of geospatial data, for example, such confounding variables include factors such as time, geographic location, and socioeconomic status information, which affect both the invariant and background representation variables, resulting in a spurious correlation. This is when \(P_{\textit{tr}}(Y=y \mid C=c, S=s) > P_{\textit{tr}}(Y=y \mid C=c)\), which causes the utilized model to learn the background representations for predicting labels. However, when the environment/selection bias changes, the distribution (features/structure) shifts, and the background representation changes. At this point, \(P_{te}(Y=y \mid C=c, S=s) < P_{te}(Y=y \mid C=c)\), which results in the original predictive relationships based on background representations no longer being applicable.

Recent studies have shown that the underlying causal mechanisms are invariant across domains\cite{32,ligr,inrl}. Therefore, causa-\\lity-based models are particularly effective at solving the OOD generalisation problem\cite{31graphood,33,34,35}. For the problem concerning feature distribution shifts, invariant representations that are causally related to labels can be learned by considering structural causal models or by removing the spurious correlation between the representations. However, considering the complexity of geographic networks lies in the diversity and instability of relationships between nodes. Specifically, nodes in geographic networks exhibit the following characteristics: a) a large number of neighbours; b) unstable/uncertain neighbour properties. Such features mean that when acquiring node representations, if neighbours are not selectively filtered, noise introduced by neighbours can affect the model's generalization ability. Therefore, in geographic networks, in addition to the target node features, the first-order neighbourhood also contains important information that can affect the subsequent representation learning and prediction tasks\cite{36good,37,38}. For example, the first-order neighbourhood of a target node has similar features/structures influenced by the environment or selection bias, and the model amplifies such similar features/structures and overrelies on similar features and structures. Therefore, when the structure distribution changes, the prediction results decrease. Therefore, a background representation change can be caused by both a feature distribution shift and a structural distribution shift of the target node. Here, we refer to the phenomenon of changes in the proportional distributions of heterogeneous and homogeneous neighbour nodes as structural distribution shifts. However, it is not clear how the invariant representations in structural distribution shifts should be learned.

In this work, we synthesise structural distribution shifts and feature distribution shifts and propose a feature-structure mixed invariant representation learning (FSM-IRL) model by combining causal theory knowledge and graph neural network techniques. To address the structural distribution shift problem, we introduce a sampling method based on causal attention, which assigns different sampling weight values to similar and dissimilar nodes. (1) For similar nodes, the causal bootstrapping method is used to estimate the causal effect between similar neighbour nodes and labels; (2) for dissimilar nodes, a single-head attention mechanism is used to assess the importance of dissimilar neighbours. Sampling weights are obtained through this method, and nodes with higher weight coefficients are selected for aggregation as a way to strengthen the ability of the model to identify invariant representations and reduce the reliance on background representations under the effect of chance or specific environmental/selection biases. To mitigate the feature distribution shift problem, the HSIC method is applied to adjust the sample weights with the aims of ensuring that all representation dimensions are independent and eliminating the spurious correlations between them; this is done as a way to allow the model to focus on invariant representations that have more stable relationships with the labels.

In summary, our contributions in this paper are as follows.

\begin{enumerate}[\hspace{1em}1.]
 \item The FSM-IRL method, which integrates structural and feature distribution shifts and achieves OOD generalisation by both enhancing invariant representations and removing spurious correlations, is proposed.
 
 \item An innovative causal attention-based sampling approach is introduced to address structural distribution shifts. This approach integrates an attention mechanism with causal bootstrapping, which aims to optimise the node sampling process and reduce the dependence of the model on background representations.
 
 \item For feature distribution shifts, this paper introduces a sample reweighting decorrelation strategy based on the HSIC method. This strategy encourages the model to eliminate the spurious correlations between representations, ignore background representations, and learn invariant representations to improve the generalizability of the model.

 \item To validate the effectiveness of the method, we create a geographic node classification dataset and design geogra-\\phic network and social network datasets as OOD datasets. We conduct experiments on unbiased datasets with different levels of bias (Cora, Pubmed, Citeseer, GOOD-Twitch, GOOD-Arxiv, and Education). Our model achieve maximum improvements of 21.07\%, 9.53\%, 5.44\%, 23.5\\6\%, 12.52\%, and 47.32\% over the baseline models, demo-\\nstrating the effectiveness of our approach.
\end{enumerate}

\section{Related works}
\label{section2}

In recent years, graph neural networks have been widely researched\cite{39}. GCN\cite{40gcn}, which is an important breakthrough in this field, utilizes a method to aggregate node features from first-order neighbourhoods. Based on this, the SGCN\cite{41} and FastGCN\cite{42} improve the computational efficiency and scalability of GCNs by optimising the computational process and sampling strategy. A graph attention network (GAT)\cite{43gat} provides greater flexibility for node feature aggregation by designing an attention mechanism. In addition, GraphSAGE\cite{44graphsage} utilizes a sampling and aggregation strategy that allows the constructed model to efficiently learn node representations in large-scale graphs. 

\sloppy
However, graph neural networks still face challenges in addressing distribution shifts in geographic networks. This problem has been widely recognised as a key topic in geographic network research. For example, GDOT\cite{otgnn} proposed a strategy that introduces Optimal Transport (OT) techniques to quantify and minimize the cost of the feature and label distribution shifts between the source and target domains, thus optimizing the domain adaptation of graph models. According to surveys, the current mainstream solution strategies are mainly classified into three categories, data-based approac-hes\cite{45,46,47,48,49,50}, model-based approaches\cite{51,52,53}, and learning str-ategy-based approaches\cite{54,55,56}, each of which has its own uni-que application scenarios and effects.

Recent research has shown that a strategy for learning geographically invariant representations using the principle of causal invariance is particularly effective when faced with the problem of performance degradation due to feature distribution shifts\cite{inrl}. Initial methods such as IRM, SDRO, and AIL have demonstrated through theory and experiments the importance of learning invariant relationships in changing environments\cite{irm,srdo,ail}. This strategy works to learn representations whose relationships with labels remain invariant despite environmental changes. In practice, graph models often overrely on spurious correlations in the input training data, leading to sharp performance degradations in the face of distribution shifts (i.e., OOD, out-of-distribution) in test data. In contrast, graph models based on causal inference attain improved generalisability in OOD scenarios by capturing the causal relationships between the input graph data and the labels to stabilize the relationships under different distributions. We can classify approaches into three types from the perspective of causal invariance: a) methods based on confounder balancing\cite{57dgnn,58oodgnn,59cmrl,60stablegnn}, b) methods based on structural causal models\cite{61dse,62cal,63ciga,64size,65gMPNNs}, and c) methods based on counterfactual inference\cite{66,67}.

\textbf{Methods based on confounder balancing.} A debiased GNN (DGNN) framework was proposed to address the OOD problem in node classification tasks\cite{57dgnn}. It uses a differential decorrelation regulariser to obtain sample weights and eliminate spurious correlations in the representation dimensions. The stability of the predictions obtained for unknown test nodes is then improved by sample weights. OODGNN\cite{58oodgnn} was developed with the belief the dependency between representations is the main reason for the degradation of model performance in OOD settings. To eliminate the statistical dependencies between all representations of a graph encoder, a nonlinear graph representation of random Fourier features was used to decorrelate the method and eliminate spurious correlations. Considering that the sample reweighting approach is computationally expensive in cases with large-scale graphs, the authors proposed a scalable global local weight calculator to learn the sample weights of each graph for optimisation step. By detecting core substructures that are causally related to chemical reactions, CMRL\cite{59cmrl} is robust to distributional variations when learning molecular relationships. The researchers first formed a hypothesis concerning causality based on molecular science domain knowledge and constructed a structural causal model (SCM) that revealed the relationships between variables. Based on the SCM, a novel conditional intervention framework, whose intervention strategy was conditioned on paired molecules, was introduced. This approach successfully mitigates the confounding effect of shortcut substructures that are spuriously correlated with chemical reactions. From the subgraph perspective, StableGNN\cite{60stablegnn} considers each subgraph to be either an irrelevant feature or a relevant feature, and the correlation between the two types of features gives rise to relationships between the irrelevant features and spurious correlations between categorical labels. The argued that spurious correlations will lead to model performance degradations after the environment changes. To eliminate the effect of subgraph-level spurious correlations, the problem was analysed from a causal perspective, and a stable GNN-based causal representation framework was proposed. The developed model employs a graph pooling layer in an end-to-end manner to extract high-dimensional representations based on subgraphs. Additionally, inspired by the confounder balancing approach for causal inference, causal variables are used to distinguish regularisers and decorrelate the data by learning sample weights. These articles investigated the removal of feature correlations in neural networks and made important contributions to improving the robustness of GNNs in the face of unknown data and environmental changes.

\textbf{Methods based on structural causal models.} According to DSE\cite{61dse}, the OOD effect acts as a confounder and is the main cause of the spurious correlations between the importance levels of subgraphs and model predictions. For this reason, researchers proposed deconfounded subgraph evaluation (DSE) method to assess the causal effects of explanatory subgraphs on model predictions. Moreover, models typically use noncausal features for prediction. However, if the given data distribution is inconsistent, for example, when noncausal features are missing in the test set, the generalizability of the model will be affected at this point. Therefore, to make the model independent of shortcut features, the relationships between the causal features and predictions can be correctly learned, and causal attention learning (CAL)\cite{62cal} was thus proposed. The CAL method draws on backdoor adjustments from causal theory, which mitigate the effect of confounding by cutting off the backdoor path using the do-calculus. To achieve graph invariance, a new framework for causality-inspired invariant graph learning (CIGA)\cite{63ciga} was proposed. The authors found that generalisation could be achieved when using a causal model to focus attention on the subgraphs that contain the most information about the cause of a label. Therefore, this article proposed an information-theoretic goal to maximise the retention of invariant intraclass information to ensure that their model was not affected by distribution shifts.
\fussy

\begin{figure*}[t]
\centering
\includegraphics[width=\textwidth]{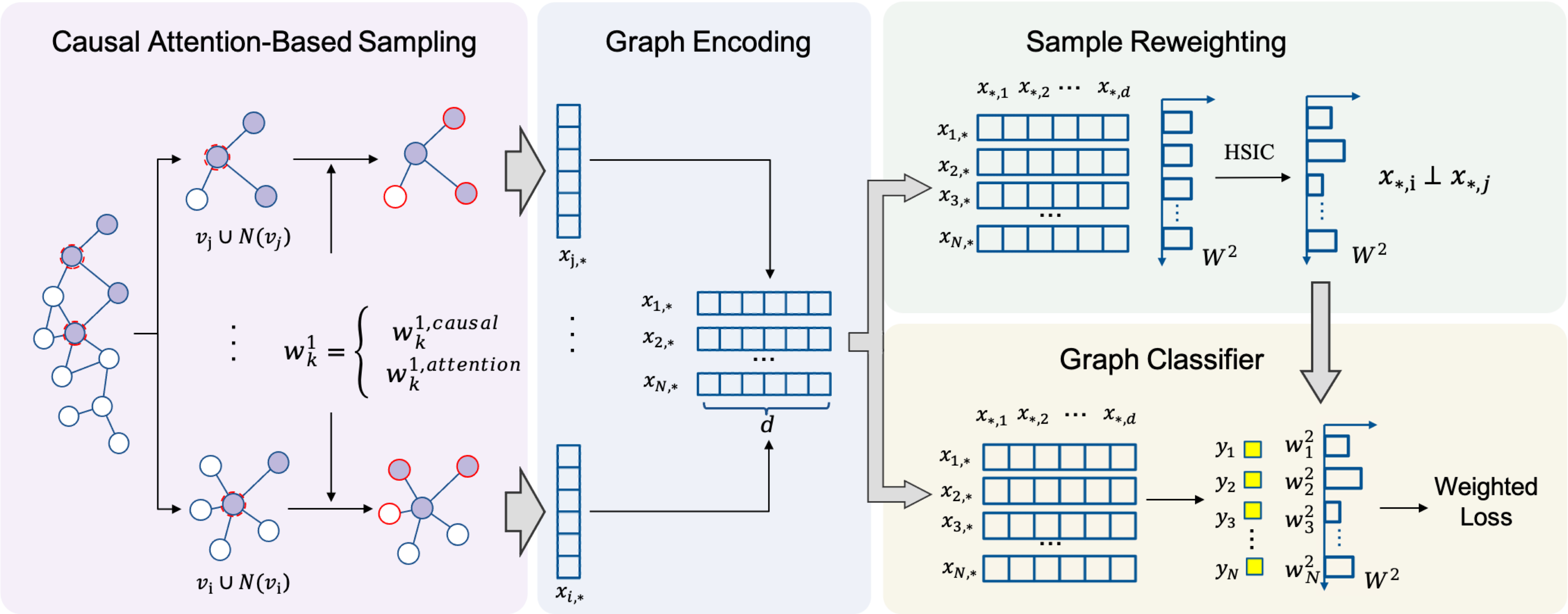}
\caption{An overview of the FSM-IRL framework. }
\label{fig1}
\end{figure*}

\textbf{Methods based on counterfactual inference.} CFLP\cite{66} starts from the counterfactual question "If the structure of a graph were different from the observations, would the link still exist?" It has been argued that counterfactual links will be able to augment graph data for representation learning purposes. Th-erefore, the associated article used a causal model to create counterfactual links and learn representations from observed and counterfactual links. CAF\cite{67} investigates the fair graph learning problem from a causal perspective, and the authors proposed a new CAF framework. This framework can select counterfactuals from training data as a way to learn fair node representations for node classification tasks.

All of the above methods for learning invariant representations based on causal learning explore the OOD problem in terms of the input features. However, graph data are topologically complex, with irregularities and connectivity. Therefore, we should not only consider feature distribution shifts but also focus on the impacts of structural distribution shifts.

\section{Method}
\label{section3}

In this paper, a hybrid feature-structure invariant representation learning model is proposed, as shown below. Overall, the model consists of 2 processes.

First, to address the effects of structural distribution shifts due to the unique structure of geographic networks, we propose a causal attention-based sampling method. This approach assigns weights to similar and dissimilar neighbours by applying different strategies, giving preference to 1) nodes among similar neighbours that have stronger causal effects on the labels and 2) nodes among dissimilar neighbours that are more similar to the target node to reduce the introduction of noise and reduce the dependence of the model on the background representation. Second, starting from the feature distribution shift aspect, the HSIC is used to achieve sample reweighting to eliminate spurious correlations among the representations and mitigate the learning of background representations.

\subsection{Preliminaries}
The graph dataset is $G = (V, E)$, $V = \left\{ v_1, v_2, \ldots, v_N \right\}$, which denotes the nodes in the graph; the number of nodes is $N=|V|$, $E$ denotes the edges in the graph, and the neighbours of node $v_i$ are denoted as $N(v_i)$. Each node $v_i$ has a corresponding label $y_i$ and its sample space is $\Omega Y$. The representation space of node $V$ is $X \in \mathbb{R}^{N^*d}$ and d is the dimensionality of the representation, where $X_{i*}$ denotes the representation of the $i$th node $v_i$ out of $N$ nodes, and $X_{*j}$ denotes the $j$th dimensional representation out of $d$ dimensions. The representations can be grouped into two types: $X = (C, S)$, where $C$ stands for invariant representations, $S$ stands for background representations, and invariant representations $C$ and background representations $S$ produce spurious correlations due to the presence of confounding variables. $W^{1} = (w_{1}^{1}, w_{2}^{1}, \ldots, w_{N}^{1})$ denotes the weights used when sampling the neighbours of the graphical model. $W^{2} = (w_{1}^{2}, w_{2}^{2}, \ldots, w_{N}^{2})$ represents the sample loss weights, where $w_{i}^{2}$ is the loss weight of the $i$th node $v_i$ in $G$. By optimally obtaining the sampling weights $W^{1}$ and loss weights $W^{2}$, a robust representation is obtained, which subsequently removes the statistical dependencies of all the dimensions on the representation $X$ and achieves the purpose of removing the spurious correlations between the invariant and background representations. This leads to satisfactory generalisation performance when predicting labels.

\subsection{Causal Attention-Based Sampling Method}
In this subsection, we introduce a new sampling method that combines an attention mechanism and causal theory, aiming to obtain sampling weights for neighbours belonging to the same category and neighbours of different categories, respectively. The aim is to allow the model to identify nodes that have strong causal relationships with a label or to identify nodes that are more similar to the target node. In this way, invariant representations are made salient, and the dependence on background representations (by chance or in specific patterns) is weakened.

\begin{figure}[t]
\centering
\includegraphics[width=0.3\textwidth]{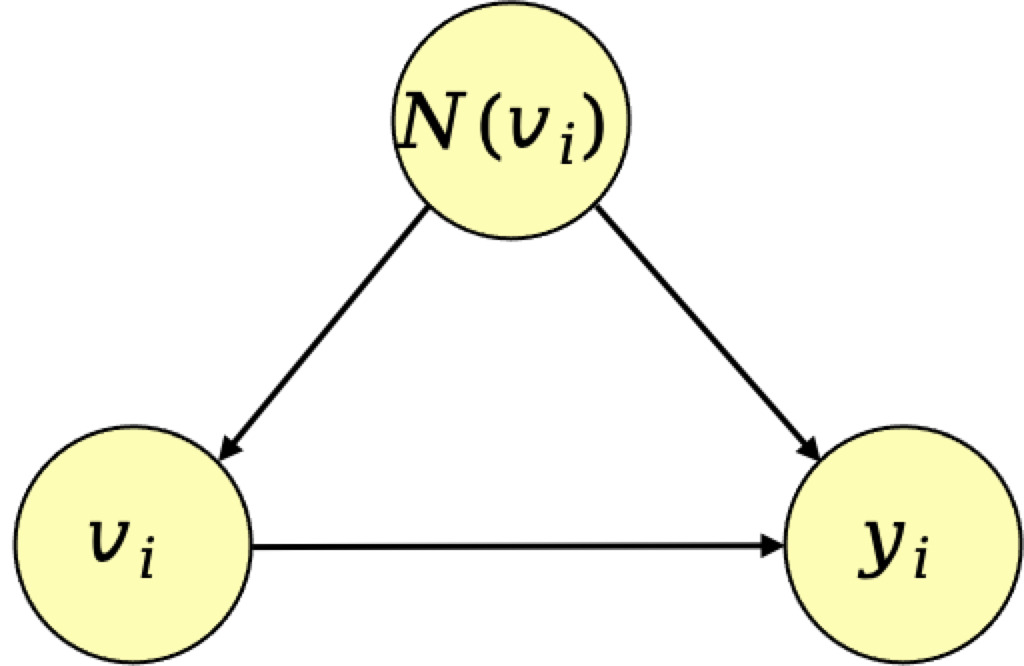}
\caption{SCM.  }
\label{fig2}
\end{figure}

\textbf{Homogeneous neighbour sampling.} Node $v_n$ and its same-class neighboring node $v_i$ have the same label value, i.e., $y_{v_n} = y_{v_i}$. Therefore, by obtaining the causal effect value from $v_i$ to $y_{v_i}$ (which we simplify as $y_i$ below), we can obtain the causal weight values of all neighboring nodes of $v_n$ in order to assess the impact of $v_i$ on the prediction of $y_n$ .

At the node level, when the GNN learns a geographic network representation bas-ed on a message propagation mechanism, the target node $v_i$ is affected by its neighbourhood $N(v_i)$. At this point, a causal path is present between $v_i$ and $N(v_i)$: ${v}_i \to N({v}_i)$. Moreover, from the classification task of the GNN, any node $v_i$ and its neighbourhood $N(v_i)$ are used to predict its label $y_i$, which indicates that there is always a causal arrow from ${v}_i \cup N({v}_i)$ to $y_i$.

By analysing this problem, we develop a structural causal model (SCM), as shown in Figure \ref{fig2}.

By constructing the SCM, we find that in addition to the causal path of the target node pointing to the label, a spurious correlation is present between the two, which is caused by the neighbouring node $N(v_i)$. In the following, we use $U$ to refer to $N(v_i)$. This spurious correlation interferes with the direct causal path between $v_i$ and $y_i$, making it more difficult for $v_i$ to estimate $y_i$. The causal graph is known to satisfy the following conditions:

1) No offspring of $v_i$ are contained in $U$.

2) $U$ blocks all causal paths from $v_i$ to $y_i$\cite{68,69}.

Thus, to identify the direct causality between $v_i$ and $y_i$, we use the backdoor criterion to exclude all noncausal pathways. This process applies the do-operator to calculate the conditional distribution \( p(y_i \mid do(v_i)) \)\cite{80cgraphsage,bootstrapping}, which is:

\begin{equation}
p(y_i|\text{do}(v_i)) = \int p(y_i|v_i, U)p(U) \, dU
\end{equation}

Then, the intervention distribution is obtained using the kernel density estimate (KDE) $ K[ \cdot ] $ derived from the joint and edge reproduction kernel Hilbert space (RKHS) functions.

\begin{equation}
p(y_i|\text{do}(v_i)) \approx \frac{1}{N} \sum_{n \in \mathbb{N}} K\left[y_i - y_n\right] \frac{K\left[v_i - v_n\right]}{\hat{p}(v_i|U)}
\end{equation}

The choice of the RKHS kernel function $K[\cdot]$ depends heavily on the sample space of the variables; here, $\Omega Y$ is discrete, and for the kernel function, we use the Kronecker delta function when addressing discrete variables.

From the above equation, we define the causal weight of $ v_i $ as follows:

\begin{equation}
w_{i}^{\text{causal}} = \frac{1}{N\hat{p}(v_i|U)}
\end{equation}

At this point, the estimated intervention distribution $ p(y_i \mid do(v_i)) $ is expressed as shown below:

\begin{equation}
p(y_i|\text{do}(v_i)) \approx w_{i}^{\text{causal}} \sum_{n \in N} K[y_i - y_n]K[v_i - v_n]
\end{equation}

Based on this expression, we can obtain the causal sampling weighting rules for the node $ v_i $ pointing to label $ y_j $:

\begin{equation}
p(y_i \mid \text{do}(v_i)) \approx 
\begin{cases} 
w_{i}^{\text{causal}} & v_i = v_n , y_i = y_n \\
0 & \text{other} 
\end{cases}
\end{equation}

The above equation implies the approximate causal weight value of the intervention distribution when the neighbour node $v_n$ points to the label $y_n (y_n=y_i)$ if the neighbour node label $y_n$ and the target node label $y_i$ are the same. However, when the neighbour node label $y_n$ and the target node label $y_i$ are different, the intervention distribution results in 0 if node $v_n$ points to label value $y_n (y_n \neq y_i)$.

\begin{table*}[ht]
\begin{center}
\label{algorithm}
\resizebox{\textwidth}{!}{
\begin{tabular}{l}
\hline
\textbf{FSM-IRL: first-order neighborhood causal attention-based sampling} \\
\hline
\textbf{Input:} $G = (V, E), \, v \in V, \, n \in \{1, \dots, N\},$ perturbation variable set $U = pa(y_i) \backslash v_i,$ causal attention-based sampling number $s = Z^+$ \\
\textbf{Output:} The set $CA = \{v_1, \dots, v_s\}$ \\
1. \ \ \  \textbf{for} $n = 1, \dots, N$ \textbf{do} \\
2. \ \ \ \ \ \ \ Find the first-order neighborhood of $v_n: N(v_n)$ \\
3. \ \ \ \ \ \ \ \textbf{for} each $v_i \in N(v_n)$ \textbf{do} \\
4. \ \ \ \ \ \ \ \ \ \ \ \textbf{if} $y_i = y_n$ \textbf{then} \\
5. \ \ \ \ \ \ \ \ \ \ \ \ \ \ \ Calculate KDEs for joint PDF of $v_i$ and $U$: $\hat{p}(v_i, U) \leftarrow \frac{1}{N} \sum_{n \in N} K[v_i - v_n ] K[U - U_n]$ \\
6. \ \ \ \ \ \ \ \ \ \ \ \ \ \ \  Calculate KDEs for PDF of $U$: $\hat{p}(U) \leftarrow \frac{1}{N} \sum_{n \in N} K[U - U_n$] \\
7. \ \ \ \ \ \ \ \ \ \ \ \ \ \ \  Obtain conditional PDF of $U$: $\hat{p}(v_i | U) \leftarrow \frac{\hat{p}(v_i, U)}{\hat{p}(U)}$ \\
8. \ \ \ \ \ \ \ \ \ \ \ \ \ \ \  Obtain the causal weight $w_i^{causal}$: $w_i^{causal} \leftarrow \frac{1}{N \hat{p}(v_i | U)}$ \\
9. \ \ \ \ \ \ \ \ \ \ \ \textbf{end if} \\
10. \ \ \ \ \ \ \ \ \ \textbf{elif} $y_i \neq y_n$ \textbf{then} \\
11. \ \ \ \ \ \ \ \ \ \ \ \ \ Kaiming uniform initialization gets $W$ \\
12. \ \ \ \ \ \ \ \ \ \ \ \ \ Obtain the attention weight $w_i^{attention}$: $w_i^{attention} \leftarrow \text{softmax}(W \cdot [x_i, x_n])$ \\
13. \ \ \ \ \ \ \ \ \ \textbf{end if} \\
14. \ \ \ \ \ \textbf{end for} \\
15. \  \textbf{end for} \\
16. \ Sample nodes in $N(v_n)$ according to $w_i$: $CA = \{v_1, \dots, v_s\}$ \\
where $K[\cdot]$ is the discrete Kronecker delta, that is $K[\cdot] = 1[\cdot]$, and PDF is the Probability Density Function. \\
\hline
\end{tabular}}
\end{center}
\end{table*}

\textbf{Heterogeneous neighbour sampling.} For the heterogeneous neighboring node $v_i$ of $v_n$, although the fact that the labels of the heterogeneous neighbour nodes do not coincide with the label of the target node, the nodes that are neighbours still have some influence on the target node. To reduce the noise or irrelevant information that may be introduced during the learning process, we obtain the sampling weights of these heterogeneous neighbour nodes using a single-head attention mechanism:

\begin{equation}
w_{ni}^{\text{attention}} = \text{softmax}(\mathbf{W} \cdot [\mathbf{x}_n; \mathbf{x}_i])
\end{equation}

$x_n$ and $x_i$ are the features of target node $v_n$ and its neighbour $v_i$, respectively; $W$ is the weight matrix in the fully connected layer; and the softmax function normalises the calculation results. With this strategy, we can preferentially select the heterogeneous neighbour nodes that are more similar to the target node in terms of their features, thus improving the accuracy and efficiency of the information aggregation process.

Combining the sampling method for $v_n$'s neighbor $v_i$, we obtain the following sampling rule:

\begin{equation}
W_{i}^{1} = 
\begin{cases} 
w_{i}^{\text{causal}} & \text{if } v_i = v_n \text{ and } y_i = y_n \\
w_{i}^{\text{attention}} & \text{otherwise}
\end{cases}
\end{equation}

We call $W_{i}^{1}$ the causal attention weight, replace uniform random sampling with weighted sampling, and refer to this process as causal attention sampling.

\subsection{Sample Reweighting Method for Representation Independence}

To address the occurrence of shifts in the structural distribution of a geographic network, we propose a causal attention-based sampling approach, which aims to reduce the overreliance of the constructed model on structural features (background representations) and to improve its adaptability when structural distributions are shifted. However, in addition to structural distribution shifts, feature distribution shifts exhibited by the geographic network target nodes themselves also pose an important challenge. Therefore, we propose that a further distinction between invariant and background representations is needed to enable the model to more effectively identify invariant representations.

To address this problem, we propose representation-indepe-ndent sample reweighting methods inspired by OODGNN\cite{58oodgnn}. The aim is to remove the statistical dependence of all dimensions in the node representations, \(X_{*,i} \perp \!\!\! \perp X_{*,j}, \quad \forall i, j \in [1, d], i  \neq j\), to achieve independence among the representation variables and to remove the spurious correlations between the invariant and background representations, such as \(C \leftrightarrow S\) in Figure \ref{fig2}. Specifically, we introduce the Hilbert–Schmidt independence criterion (HSIC)\cite{70,71} to achieve independence among the representation variables. The independence of different dimensions is measured based on the HSIC metric:

\begin{equation}
\text{HSIC}(X_{*i}, X_{*j}) = 0 \iff X_{*i} \perp \!\!\! \perp X_{*j}
\end{equation}

First, we express the HSIC in matrix form in a case with finite samples. which is expressed as follows:

\begin{equation}
\begin{aligned}
\text{HSIC}(X_{*i}, X_{*j}) &= \frac{1}{n^2} \text{Tr}(k_{x_{*i}} k_{x_{*j}}) + \frac{1}{n^4} \text{Tr}(k_{x_{*i}} 1k_{x_{*j}} 1) \\ & - \frac{2}{n^3} \text{Tr}(k_{x_{*i}} k_{x_{*j}} 1) = \frac{1}{n^2} \text{Tr}(k_{x_{*i}} Jk_{x_{*j}} J)
\end{aligned}
\end{equation}


where \(k_{x_{*i}}\), \(k_{x_{*j}}\) are $n*n$ symmetric matrices, $Tr$ denotes the trace of a matrix, and $1$ denotes a full matrix of $1s$ with a size of $n \times n$. Here, $J = I-1/n$, and $I$ is an $n$th-order unit matrix. However, this is a biased estimate, and replacing the previous $1/n$ with $1/(n-1)$ yields an unbiased estimate:

\begin{equation}
\text{HSIC}(X_{*i}, X_{*j}) = \frac{1}{(n - 1)^2} \text{Tr}(k_{x_{*i}} Jk_{x_{*j}} J)
\end{equation}

\sloppy
Second, based on the above independence criterion, we use reweighting to achieve independence for the variables in the representation. Sample reweighting is one of the traditional RCM\cite{72} methods, and its core idea lies in the fact that the distributions of the confounding variables are consistent between the treatment and control groups through the assignment of different weights to samples with different characteristics\cite{73}. That is, eliminating the confounding variables ensures a randomised trial. Specifically, we incorporate weights \(W^2 = (w_1^2, w_2^2, \ldots, w_N^2)\)  into the HISC calculation in the above equation:

\begin{equation}
\text{HSIC}(W^2 X_{*i}, W^2 X_{*j}) = \frac{1}{(n - 1)^2} \text{Tr}(k_{W^2 X_{*i}} J k_{W^2 X_{*j}} J)
\end{equation}

Finally, the dependencies between representation dimensions are minimised by minimising the HISC. Therefore, weight values are obtained:
\fussy

\begin{equation}
W^2_{\text{min}} = \underset{W}{\text{argmin}} \sum_{1 \leq i < j \leq d} \text{HSIC}(W^2 X_{*i}, W^2 X_{*j})
\end{equation}

During the process of optimising the graph weights \(W^2\), iteratively minimising the HISC results expressed in matrix form in the above equation removes as much dependency between the representations as possible.

\section{Experiments and results}
\label{section4}

\subsection{Experimental Setup}
\subsubsection{Datasets}
\sloppy
We build bias datasets for social and geographic networks based on the notion presented in this paper regarding the coexistence of feature and structural distribution shifts. The summary statistics of the datasets are shown in Table \ref{tab1}. Cora, Pubmed\cite{74pubmed}, and Citeseer\cite{75citeseer}, which are widely used benchmark datasets, have been strongly recognized in the academic community, which makes our study easy to compare with other works. Similarly, we consider the GOOD-Twitch and GOOD-Arxiv\cite{36good} large-scale datasets. Education is a geographic domain dataset and has different characteristics than social network datasets, such as spatial relationships and geographic attributes. Choosing these datasets for our experiments helps us explore the effectiveness of our method in terms of addressing the OOD problem in geographic networks while also providing insights into its performance across different types of networks.

\begin{enumerate}

\item \textbf{Geographic network datasets:}

\textbf{Education:} In this paper, we build a node-disaggregated educational attainment dataset for 2021 at the county level in the United States. We use the U.S. county level as the nodes and obtain statistics for 2021 through the U.S. Census American Community Survey (ACS), which includes education, economy, poverty rate, race and other statistics, as the attribute information. Xindong Li and Xiang Zhao start from the individual level, converting educational tiers into corresponding years of educational attainment to define education levels. We construct labels based on this definition of “education level”. In this paper, we obtain the average number of years of education earned by the population in each county-level region, and four types of labels are produced after the division step. A dataset concerning the multiscale dynamic movement of people in the U.S. during the COVID-19 epidemic period is used to construct a movement network of people for the whole year of 2021\cite{76,77,78}, and county-level neighbourhoods are established as connecting edges. A total of 809 nodes, 593,996 edges, 218 dimensional features, and four categories are finally obtained for the education level dataset.

\item \textbf{Social network datasets:}

\textbf{Cora:} The Cora dataset is a commonly used dataset for citation network classification tasks in the graph data domain. It contains an academic paper citation network covering papers in the field of machine learning, with 2708 papers, 5429 edges, and 1433 feature dimensions across 7 categories.

\textbf{Citeseer:} The Citeseer dataset is a dataset that is widely used in academic citation network classification tasks. The dataset contains citation networks of papers in the field of computer science, with 3,312 papers, 4,723 edges, and 3,703 feature dimensions, which cover 6 categories.

\textbf{Pubmed:} The Pubmed dataset is a commonly used dataset for literature classification tasks, especially in the biomedical field. This dataset contains 19717 abstracts from the PubMed database, 44338 edges, and 500 feature dimensions, which cover 3 categories.

\textbf{GOOD-Twitch:} This is a game player network dataset. The nodes represent players with gaming characteristics, and the edges represent friendship relationships between players. The main binary classification task of this dataset is to predict whether a player streams mature content. The GOOD-Twitch dataset is segmented by user language to ensure that the prediction targets are not affected by the language used by the user.

\textbf{Good-Arxiv:} This dataset is based on OGB's adaptation of the citation network of arXiv papers in computer science. In this directed graph dataset, nodes represent computer science (CS) papers on arXiv, while directed edges represent citation relationships between papers. Its task is to classify the subject areas of CS papers, involving 40 categories. The dataset is split according to the time domain.
\end{enumerate}
\fussy

\begin{table}[h]
\centering
\caption{Summary of the datasets used in the experiments.}
\label{tab1}
{\small
\begin{tabularx}{\columnwidth}{>{\centering\arraybackslash}p{1.9cm}>{\centering\arraybackslash}p{0.8cm}>{\centering\arraybackslash}p{0.8cm}>{\centering\arraybackslash}p{0.9cm}>{\centering\arraybackslash}p{0.9cm}>{\centering\arraybackslash}p{0.8cm}}
\hline
\textbf{Dataset} & \textbf{Graphs} & \textbf{Nodes} & \textbf{Edges} & \textbf{Features} & \textbf{Classes} \\ \hline
Education        & 1               & 809            & 593996         & 218               & 4                \\ 
Cora             & 1               & 2708           & 5429           & 1433              & 7                \\ 
Citeseer         & 1               & 3327           & 4732           & 3703              & 6                \\ 
Pubmed           & 1               & 19717          & 44338          & 500               & 3                \\  
GOOD-Twitch      & 1               & 34120          & 892346         & 128               & 2                \\
GOOD-Arxiv      & 1               & 169343          & 1166243         & 128               & 40                \\ \hline
\end{tabularx}
}
\end{table}

For the Education dataset, we construct structural offsets by perturbing the original graph structure. We plan to randomly remove 50\% of the edges of the original graph to create a new graph structure. To fully evaluate the model, we set up structural offsets: we conduct training on the original graph and then choose the graph structure with 50\% of the contiguous edges removed as the test structure. In social networks Cora, Pubmed, and Citeseer, for the training dataset \(G_{\text{train}}\), we want the model to sample nodes distributed at the edges of the categories, and for the test set \(G_{\text{test}}\), we want it to sample nodes distributed inside the categories to ensure distribution shifts. Therefore, we use the original unbiased setting (no bias) in DGNN\cite{57dgnn} and the three levels of distribution shift (small, medium, big) for each dataset (Here, the no bias, small, medium, and big shifts correspond to the unbiased, slight, medium, and heavy shift settings in the DGNN paper, respectively). We progressively increase the shift levels to validate the model's generalization ability.

\subsubsection{Baseline}
For the node classification task conducted in this paper, we choose Planetoid\cite{79planetoid}, a GCN\cite{40gcn}, GraphSAGE\cite{44graphsage}, IRM\cite{irm}, SRDO\cite{srdo}, DANN\cite{DANN}, Deep CORAL\cite{deepcoral}, and the more recent C-GraphSAGE\cite{80cgraphsage} model as baselines for comparison. Pl-\\anetoid, the GCN, and GraphSAGE are all widely recognised and used methods within the field of graph neural networks, IRM and SRDO are methods for stable learning in domain-specific data. DANN and Deep CORAL employ adversarial training and statistical alignment approaches, respectively. C-GraphSAGE is a graph neural network model that incorporates causal inference. They each represent different mechanisms for processing graph data contained in graph neural networks, thus providing solid performance baselines for our research. By using these models as baselines, we can comprehensively evaluate and demonstrate the performance of our approach. The models are specifically described below.

\begin{figure*}[t]
\centering
\includegraphics[width=\textwidth]{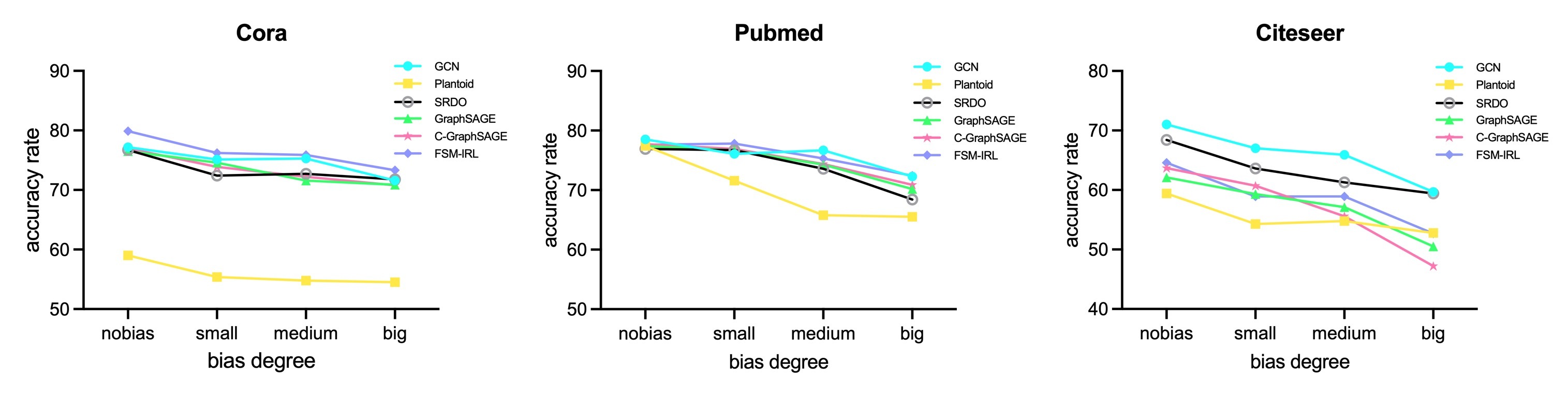}
\caption{Prediction results obtained by different models after setting the Cora, Pubmed and Citeseer datasets to different bias levels. }
\label{fig3}
\end{figure*}

\begin{enumerate}
    \item \textbf{GCN:} The GCN updates the representation of a node by applying a convolution operation on its neighbouring nodes to efficiently capture the relationships between nodes.
    \item \textbf{Planetoid:} This is a classic semisupervised graph embedding method that learns node representations for effectively performing graph node classification by combining the neighbourhood information of different nodes. In this paper, we use the inductive variant of Planetoid.
    \item \textbf{GraphSAGE:} GraphSAGE samples neighbouring nodes and aggregates their features to learn a representation of the target node and uses random wandering or sampling methods to obtain neighbourhood information, which can be applied to large-scale graph data to efficiently capture the local structural features of nodes.
    \item \textbf{C-GraphSAGE:} C-GraphSAGE operates based on the causal bootstrapping method to capture the weights between the neighbours of the target node and its labels and samples the neighbouring nodes based on their weights to compensate for the sampling bias and information loss caused by the random sampling in the GraphSAGE process, thus improving the robustness of classification.
    \item \textbf{IRM:} IRM is an invariant learning method that aims to achieve robust generalisation capabilities across different environments. This method learns environment-indepen-\\dent properties by finding data representations or features and ensuring that the best classifiers trained on these representations are adapted to all environments.
    \item \textbf{SRDO:} SRDO focuses on problems in linear predictive model learning, especially in the presence of model error setting bias. When collinearity between input variables is present, it can amplify the error in parameter estimation, leading to instability in prediction results when the training and testing distributions differ. To address this, SRDO proposes a sample reweighting method aimed at reducing collinearity among input variables.
    \item \textbf{DANN:} By introducing an additional Domain Classifier, the model simultaneously reduces the distribution discrepancy between the source and target domains while learning feature representations. This process can be vie-\\wed as adversarial training.
    \item \textbf{Deep CORAL:} By minimizing the distance between the second-order statistics of the source domain features and the target domain features, the impact of domain shift is mitigated.
\end{enumerate}

\begin{table*}[h]
\begin{center}
\caption{Prediction results on different models after setting Covariate shift on Education and Time domains for GOOD-Twitch and GOOD-Arxiv datasets respectively.}
\label{covariate}
\begin{tabular}{ccccccccc}
\hline
\multirow{1}{*}{Covariate} & \multicolumn{4}{c}{GOOD-Twitch(Education)} & \multicolumn{4}{c}{GOOD-Arxiv(Time)} \\
\cmidrule(lr){2-5} \cmidrule(lr){6-9} 
& $\text{ID}_{\text{ID}}$  & $\text{ID}_{\text{OOD}}$ & $\text{OOD}_{\text{ID}}$ & $\text{OOD}_{\text{OOD}}$ & $\text{ID}_{\text{ID}}$  & $\text{ID}_{\text{OOD}}$ & $\text{OOD}_{\text{ID}}$ & $\text{OOD}_{\text{OOD}}$ \\ 
    Planetoid   & \textbf{62.88} & 54.99 & 51.97 & 70.26 & 28.17 & 27.17 & 28.12 & 27.33\\ 
    GCN         & 48.22 & \textbf{71.19} & 48.22 & 71.19 & 28.80 & 26.10 & 28.80 & 26.10 \\ 
    Graphsage   & 53.82 & 58.70 & \textbf{56.48} & 49.59 & \textbf{42.92} & \textbf{41.48} & \textbf{41.27} & \textbf{42.72} \\ 
    C-Graphsage & 52.05 & 37.50 & 53.35 & 37.19 & 36.37 & 37.67 & 40.70 & 35.27 \\ 
    FSM-IRL     & 55.09 & 44.68 & 54.53 & 43.65 & 33.16 & 25.95 & 32.00 & 25.53 \\ 
\hline
\end{tabular}
\end{center}
\end{table*}

\begin{table*}[h]
\begin{center}
\caption{Prediction results on different models after setting Concept shift on Education and Time domains for GOOD-Twitch and GOOD-Arxiv datasets respectively.}
\label{concept}
\begin{tabular}{ccccccccc}
\hline
\multirow{1}{*}{Concept} & \multicolumn{4}{c}{GOOD-Twitch(Education)} & \multicolumn{4}{c}{GOOD-Arxiv(Time)} \\
\cmidrule(lr){2-5} \cmidrule(lr){6-9} 
& $\text{ID}_{\text{ID}}$  & $\text{ID}_{\text{OOD}}$ & $\text{OOD}_{\text{ID}}$ & $\text{OOD}_{\text{OOD}}$ & $\text{ID}_{\text{ID}}$  & $\text{ID}_{\text{OOD}}$ & $\text{OOD}_{\text{ID}}$ & $\text{OOD}_{\text{OOD}}$ \\ 
    Planetoid & \textbf{62.89} & 55.19 & 50.82 & \textbf{62.78} & 37.68 & 23.15 & 37.69 & 23.17 \\
    GCN & 50.08 & \textbf{62.15} & 50.08 & \textbf{62.15} & 37.12 & 18.38 & 37.12 & 18.38 \\
    GraphSage & 57.48 & 51.04 & 56.54 & 49.00 & \textbf{47.65} & 36.13 & 47.22 & 34.02 \\
    C-GraphSage & 37.62 & 37.92 & 52.01 & 37.48 & 41.50 & \textbf{40.40} & 42.17 & \textbf{41.30} \\
    FSM-IRL & 61.18 & 43.06 & \textbf{58.67} & 43.21 & 38.15 & 30.90 & 33.80 & 26.08 \\
\hline
\end{tabular}
\end{center}
\end{table*}

\subsubsection{Implementation Details}
Our experiments impose uniform restrictions on the model parameters. During model training, we set the number of epochs used for the social network dataset to 20 and the batch size to 20, while for the geographic network dataset, we set the number of epochs to 110 and 150 and the batch size to 40. Additionally, for the Citeseer dataset, on all models except SRDO we set the learning rate lr to 0.1, set lr for GOOD-Arxiv to 0.001, and we set the learning rate of the other datasets to a uniform value of 0.01. We set the $L2$ regularisation constant to $1e-3$ for all datasets and use Adam optimisation to update the weights of the neural network. The neural network $K$ has two layers; the hidden-layer dimensionality d for Pubmed is 256, and that for all other datasets is set to 128. The dimensionality of the output layer is equal to the number of labels. The update period of the learning loss weights is 100 epochs. Regarding the training, validation and test sets of the social network, we follow the experimental setup of the original paper, where 20 samples from each category are taken as the training set and 1000 nodes are selected as the test set to evaluate the classification performance of the tested methods. For the geographic networks, we uniformly divide the dataset, with 70\% for the training set, 20\% for the validation set and 10\% for the test set. The experiments are run in Python 3.7 and PyTorch 1.12.1. Two deep learning stations are used to run all the experiments: a DGX A100 with 4×80 G and an A6000 with 3×48 G. We ensure in our experiments that the validation set and the training set do not follow the same distribution, and finally, the mean and standard deviation of the 3 experimental repetitions are determined.

\begin{table*}[h]
\begin{center}
\caption{On the Education dataset, we compare the predictions produced by different models before and after the structural distribution shift (accuracy±standard deviation \% and macro-F1±standard deviation \%).}
\label{tab2}
\begin{tabular}{cccccccc}
\hline
\multirow{1}{*}{Data} & Model & \multicolumn{2}{c}{original} & \multicolumn{2}{c}{de-50} & \multicolumn{2}{c}{bias} \\
\cmidrule(lr){3-4} \cmidrule(lr){5-6} \cmidrule(lr){7-8}
& & acc(\%) & Macro-f1 & acc(\%) & Macro-f1 & acc(\%) & Macro-f1 \\
\multirow{5}{*}{Education} & GCN & 30.86±12.83 & 49.21±42.61 & 30.86±12.83 & 49.21±42.61 & 30.86±12.83 & 49.21±42.61\\
                   & Planetoid & 16.05±0.00 & 8.21±0.00 & 16.05±0.00 & 8.21±0.00 & 16.05±0.00 & 8.21±0.00\\
                   & IRM & 30.86±10.76  & 20.15±16.52 & 37.86±11.86 & 31.17±17.09 & 40.74±7.41 & 39.92±3.92 \\
                   & DANN & 22.63±4.99  & 21.82±5.47 & 20.17±3.77 & 13.66±3.91 & 22.22±11.78 & 14.72±7.22 \\
                   & Deep CORAL & 31.28±3.77  & 28.01±4.67 & 18.93±1.43 & 14.09±3.27 & 22.63±6.80 & 19.03±9.71 \\
                   & GraphSAGE & 55.56±8.10 &55.28±9.65 & 40.33±22.81 & 29.56±32.61 & 50.62±10.54 & 49.50±7.61\\
                   & C-GraphSAGE & 37.45±15.53 &36.87±15.79 & 48.56±14.31 & 47.26±14.87 & 50.62±10.47 & 44.85±12.90 \\
                   & FSM-IRL & \textbf{60.49±3.49} & \textbf{57.51±2.24} & \textbf{63.37±5.23} & \textbf{60.77±3.89} & \textbf{59.67±5.14} & \textbf{52.88±5.78} \\
\hline
\end{tabular}
\end{center}
\end{table*}

\subsection{Comparative experiments on small-scale social network datasets}
To test the performance of our model, we evaluate the performance of the baselines and the proposed method. The results of comparisons conducted on the small-scale social network dataset are shown in Figure \ref{fig3}. We use open source code to implement the GCN, Planetoid, GraphSAGE, SRDO, and C-GraphSAGE models in PyTorch. Four of the models, GraphSAGE, C-GraphSAGE, SRDO, and FSM-IRL, are parameteri-\\sed under equivalent conditions, and the rest of the model hyperparameters are set to the hyperparameters reported in the original papers.

By observing Figure \ref{fig3}, we can observe that: (1) Our proposed model achieves the best performance in most cases, which is good proof of the effectiveness of our proposed FSM-IRL strategy. Compared to the baseline models, our proposed model achieves maximum performance improvements of 21.07\%, 9.5-\\3\%, and 5.44\% on the Cora, Pubmed, and Citeseer datasets, respectively. (2) A comparison of the prediction results on unbiased and biased datasets reveals that structural distribution shift affects the model's performance, leading to a decline. (3) As the shift degree increases, the performance of all models decreases; however, FSM-IRL exhibits greater robustness, resulting in more significant improvements. This suggests that FSM-IRL has better robustness on datasets with larger shifts. (4) A comprehensive comparison between our model and the other models reveals that FSM-IRL significantly outperforms the baselines on Cora, achieves similar results to those of the GCN on Pubmed, and slightly underperforms relative to the GCN and SRDO model on Citeseer. However, the results produced by FSM-IRL across all social network datasets are generally more accurate than those of the other models except the GCN.

\subsection{Comparative experiments on large-scale social network datasets}
We compared the results of GCN, Planetoid, GraphSAGE, C-GraphSAGE, and FSM-IRL on large-scale social network datasets, as shown in Table \ref{covariate} and Table \ref{concept}.

We evaluated the performance of four baseline models and the FSM-IRL model on two large-scale datasets, following covariate and concept shifts. From the experimental results we find that on the GOOD-Twitch dataset, the FSM-IRL model is relatively stable for in-distributed data, with performance degradation in the IDOOD and OODOOD scenarios. On the GOOD-Arxiv dataset, the overall performance of FSM-IRL is lower compared to GOOD-Twitch, and it generally performs better in in-distribution scenarios than in out-of-distribution tests. This suggests that the FSM-IRL model faces greater challenges when dealing with multi-category, large-scale datasets.

\subsection{Comparative experiments on geographic networks}
We evaluated GCN, Planetoid, GraphSAGE, IRM, DANN, Deep CORAL, and FSM-IRL on the geographic network dataset. The results for each model on this dataset are shown in Table \ref{tab2}.

Table \ref{tab2} shows that FSM-IRL consistently outperforms the other models in terms of prediction accuracy in all three cases. When trained and predicted on the original graph structure, FSM-IRL results in 4.93\% to 44.44\% higher accuracies than those produced by the other models. When trained and predicted on the graph structure produced after deleting the connecting edges, the accuracy of FSM-IRL is 14.81\% to 47.32\% higher than those of the other models. When trained on the original graph structure and then tested on the graph structure produced after deleting the concatenated edges, FSM-IRL achieves 9.05\% to 43.62\% better accuracy results than those of the other models. All these results demonstrate the effectiveness of the FSM-IRL model on the geographic network dataset. The results obtained from the IRM model are lower compared to FSM-IRL, demonstrating the superiority of our method over traditional invariant representation learning machine learning methods. By comparing DANN, Deep CORAL, and FSM-IRL, we find that FSM-IRL exhibits stronger robustness and superiority compared to traditional adversarial training and statistics-based domain adaptation methods. We also observe that the GCN obtains better results on the social network dataset, but on the contrary, it produces poor results on the geographic network dataset at approximately 30\%. This is because the GCN learns node representations by aggregating information about the neighbours of the target node. Social network datasets usually have very few neighbouring nodes, and the connections between their nodes are typically very strong. On the other hand, each node in our geographic network dataset has a large number of neighbours, and the neighbours contain considerable "noise", so the GCN learns considerable useless information when learning representations from the geographic network datasets. Therefore, the GCN performs better on social network datasets. Moreover, the GraphSAGE, C-GraphSAGE and FSM-IRL models, which sample neighbouring nodes to learn representations, also have higher classification accuracies than those of the GCN models. 

\begin{table*}[ht]
\begin{center}
\caption{Comparisons among the classification performances of the baseline model before and after the inclusion of the sample reweighting method under the no\_bias and three bias settings on the Cora, Pubmed and Citeseer datasets.}
\label{tab3}
\begin{tabular}{cccccc}
\hline
& & \multicolumn{4}{c}{acc(\%)} \\
\cmidrule(lr){3-6}
Data & Model & nobias & small & medium & big \\
\multirow{4}{*}{Cora} & GraphSAGE & 76.50 & 74.60 & 71.60 & \textbf{70.87} \\
                   & HSIC-GraphSAGE & \textbf{76.77} & \textbf{75.57} & \textbf{72.63} & 70.60 \\
                   & C-GraphSAGE & 76.97 & 73.87 & 72.23 & 70.77 \\
                   & HSIC-C-GraphSAGE & \textbf{77.60} & \textbf{74.43} & \textbf{73.23} & \textbf{71.27} \\
\hline       
\multirow{4}{*}{Pubmed} & GraphSAGE & 77.40 & 76.67 & 74.27 & 70.17 \\
& HSIC-GraphSAGE & \textbf{77.53} & \textbf{77.10} & \textbf{75.33} & \textbf{71.50} \\
& C-GraphSAGE & 77.70 & 76.93 & 74.37 & 70.87 \\
& HSIC-C-GraphSAGE & \textbf{77.80} & \textbf{77.77} & \textbf{74.77} & \textbf{71.97} \\
\hline       
\multirow{4}{*}{Citeseer} & GraphSAGE & 62.10 & \textbf{59.33} & 57.10 & 50.50 \\
& HSIC-GraphSAGE & \textbf{65.07} & 58.67 & \textbf{57.53} & \textbf{50.77} \\
& C-GraphSAGE & 63.67 & 60.73 & 55.57 & 47.23 \\
& HSIC-C-GraphSAGE & \textbf{66.20} & \textbf{61.93} & \textbf{56.10} & \textbf{49.77} \\
\hline
\end{tabular}
\end{center}
\end{table*}

\begin{table*}[!ht]
\begin{center}
\caption{Comparisons among the classification performances of the baseline model before and after the inclusion of the causal attention-based sampling method under the no\_bias and three bias settings on the Cora, Pubmed and Citeseer datasets.}
\label{tab4}
\begin{tabular}{cccccc}
\hline
& & \multicolumn{4}{c}{acc(\%)} \\
\cmidrule(lr){3-6}
Data & Model & nobias & small & medium & big \\
\multirow{4}{*}{Cora} & GraphSAGE & 76.50 & 74.60 & 71.60 & 70.87 \\
                   & CA-GraphSAGE & \textbf{79.27} & \textbf{76.50} & \textbf{74.80} & \textbf{73.17} \\
                   & HSIC-C-GraphSAGE & 77.60 & 74.43 & 73.23 & 71.27 \\
                   & FSM-IRL & \textbf{79.87} & \textbf{76.23} & \textbf{75.87} & \textbf{73.33} \\
\hline       
\multirow{4}{*}{Pubmed} & GraphSAGE & 77.40 & 76.67 & 74.27 & 70.17 \\
& CA-GraphSAGE & \textbf{78.10} & \textbf{77.37} & \textbf{74.73} & \textbf{71.70} \\
& HSIC-C-GraphSAGE & \textbf{77.80} & 77.77 & 74.77 & 71.97 \\
& FSM-IRL & 77.63 & \textbf{77.83} & \textbf{75.33} & \textbf{72.43} \\
\hline       
\multirow{4}{*}{Citeseer} & GraphSAGE & 62.10 & 59.33 & 57.10 & 50.50 \\
& CA-GraphSAGE & \textbf{62.93} & \textbf{60.50} & \textbf{57.10} & \textbf{52.93} \\
& HSIC-C-GraphSAGE & \textbf{66.20} & \textbf{61.93} & 56.10 & 49.77 \\
& FSM-IRL & 64.57 & 58.93 & \textbf{58.93} & \textbf{52.67} \\
\hline
\end{tabular}
\end{center}
\end{table*}

\subsection{Ablation experiments}
In this study, we use four datasets, namely, Cora, Citeseer, Pubmed, and Education. We follow uniform parameter settings and employ commonly used metrics for assessing the classification accuracies of different methods. The model accuracies achieved before and after combining our methods are compared and analysed. Specifically, Table \ref{tab3} and Table \ref{tab5} demonstrate the model accuracy changes exhibited by the baseline model before and after the introduction of the HSIC-based sample reweighting module (HSIC-) when processing the social network dataset and the geographic network dataset. Furthermore, Table \ref{tab4} and Table \ref{tab6} compare the accuracies achieved by the baseline model before and after the addition of the causal attention sampling module (CA-) when processing the same dataset. 

\sloppy
The results in Table \ref{tab3} and Table \ref{tab4} show that the accuracy of the baseline model improves after the addition of our method. Specifically, except for a few datasets where the accuracy slightly decreases after implementing the proposed strategy, basically all the baseline models show some improvement in their results after adding our method. Comparing the accuracies achieved before and after adding the sample reweighting method, the maximum improvements provided on the Cora, Pubmed, and Citeseer datasets are 1.03\%, 1.33\%, and 2.97\%, respectively, which reflect the compatibility of the sample reweighting method with various models. Comparing the accuracies achieved before and after adding the causal attention sampling method, the maximum improvements provided on the Cora, Pubmed, and Citeseer datasets are 3.20\%, 1.53\%, and 2.90\%, respectively, demonstrating the advantage of the causal attention sampling method over random sampling and causal sampling in cases with offset datasets.

\begin{table*}
\begin{center}
\caption{Comparison among the classification performances of the baseline model before and after the addition of the sample weighting method in the three cases: the original setting (original), the setting with 50\% of the edges deleted (de-50), and the original training setting but testing with the new method to achieve a distribution shift (bias).}
\label{tab5}
\begin{tabular}{ccccc}
\hline
& & \multicolumn{3}{c}{acc(\%)} \\
\cmidrule(lr){3-5}
Data & Model & original & de-50 & bias \\
\multirow{4}{*}{Education} & GraphSAGE & \textbf{55.56} & 40.33 & 50.62 \\
                   & HSIC-GraphSAGE & 50.20 & \textbf{56.79} & \textbf{56.38} \\
                   & C-GraphSAGE & 37.45 & \textbf{48.56} & 50.62 \\
                   & HSIC-C-GraphSAGE & \textbf{55.55} & 45.68 & \textbf{57.12} \\
\hline       
\end{tabular}
\end{center}
\end{table*}

\begin{table*}
\begin{center}
\caption{Comparison among the classification performances of the baseline model before and after the addition of the causal attention-based sampling method in the three cases: the original setting (original), the setting with 50\% of the edges deleted (de-50), and the original training setting but testing with the new method to achieve a distribution shift (bias).}
\label{tab6}
\begin{tabular}{ccccc}
\hline
& & \multicolumn{3}{c}{acc(\%)} \\
\cmidrule(lr){3-5}
Data & Model & original & de-50 & bias \\
\multirow{4}{*}{Education} & GraphSAGE & 55.56 & 40.33 & 50.62 \\
                   & CA-GraphSAGE & \textbf{57.61} & \textbf{60.91} & \textbf{58.43} \\
                   & HSIC-C-GraphSAGE & 55.55 & 45.68 & 57.12 \\
                   & FSM-IRL & \textbf{60.49} & \textbf{63.37} & \textbf{59.67} \\
\hline       
\end{tabular}
\end{center}
\end{table*}

In this task, we evaluate the effectiveness of the sample reweighting method and the causal attention sampling method on the Education dataset, where the structural distribution is shifted, as shown in Tables \ref{tab5} and \ref{tab6}. Comparing the results obtained before and after the addition of the sample reweighting method to the baseline model, the maximum improvement provided on Education is 18.10\%. Comparing the results obtained before and after the addition of the causal attention-based sampling method to the baseline model, the maximum improvement provided on Education is 20.58\%. These results show that on the Education dataset, our method also improves upon the original model and can attain improved generalizability.
\fussy

\section{Discussion}
\label{section5}
\subsection{Limitations}
Our work investigates the OOD problem faced by geograp-\\hic networks from a causal perspective and offers new perspectives for the analysis of geographic networks in terms of feature-structural distribution shifts. Although recent research has made significant progress in the area of geographic networks, still some unexplored issues remain.

\textbf{Failure to consider node degree bias.} Recent discussions of the structural distribution shift problem only considered the homogeneity and heterogeneity of neighbouring nodes in proportion to the variation, but the structure distribution also contains information such as the node degree. The nodal degree is an important feature of a graph structure and is crucial for understanding the structural dynamics and information propagation process of a network. Future work can focus on how to incorporate node degree bias into the geographic network analysis task and determine how it affects network behaviours and predictive models.

\textbf{Shortcomings on large-scale datasets.} There is a difference between the speeds at which our method and other models run on geographic network datasets, and we observe that a significant amount of time is consumed in the computational process when the sample reweighting module acquires weights. Future work may require the development of more efficient algorithms or the use of optimisation techniques to improve the efficiency of the model when addressing problems with large-scale dataset. Simultaneously, through our experiments, we have found that our model performs below some models on large-scale datasets such as GOOD-Arxiv, demonstrating that while our work excels on geographic network datasets, it lacks universality on other network datasets. We plan to delve into cutting-edge methods for learning invariant representations in changing environments in subsequent work to enhance the mo-\\del’s capability to capture complex data patterns\cite{ail,inrl}.

\textbf{Scarcity of geographic network datasets.} Currently, we are the first to construct a geographic network dataset, so the approach developed in this paper is only tested on the Education dataset. Future research can consider constructing more geographic network datasets to advance the research on geographic networks.

\textbf{Generalization capabilities of large models on OOD.} Fine-tuning large models is starting to gain attention across various domains, and they also face the risk of overfitting due to distribution shifts\cite{vlal}. This is a cutting-edge research area, particularly crucial in achieving out-of-distribution generalization in the face of scarce samples and imbalanced distributions. Future work could explore how to integrate our work with the large model domain to mitigate this risk.

\textbf{Consider introducing a transport cost metric.} Although our method has achieved good results in the feature-structure distribution shift problem, we have not yet explored the potential of the transport cost metric in quantifying distribution shift\cite{ot}. This is an effective metric that has shown superiority in domain adaptation and out-of-distribution generalization tasks. Future work could consider combining the transport cost metric with invariant representation learning to precisely quantify the distribution discrepancy between the source and target domains.

\subsection{Potential of causality-invariant learning}
In this paper, we incorporate causality-invariant learning into geographic representation learning to efficiently learn geographic representations under OOD settings. This approach provides a new perspective for understanding and analysing geographic data, especially for revealing the complex causal relationships between geographic features. The introduction of causal learning improves the efficiency at which geographic data can be processed and analysed and provides new insights into application areas such as urban planning, environmental monitoring, and traffic management.

\section{Conclusion and future work}
In this study, we explore the reasons why the existing models exhibit reduced generalizability when addressing geographic network distribution shifts. Through our analyses, we find that this issue mainly stems from background representation changes caused by feature distribution shifts and structural distribution shifts. We propose a novel hybrid feature-structure invariant representation learning (FSM-IRL) model to mitigate the distributional shift observed in geographic networks. The model focuses on invariant representations that have more stable relationships with labels and ignores background representations that have unstable relationships with the labels under distribution shifts to the greatest extent possible. We conduct a large number of experiments based on FSM-IRL, aiming to reveal and analyse the performance of the model in different scenarios and tasks. Through comparative studies involving other methods, we find that FSM-IRL achieves satisfactory results in terms of generalizability.

Although significant results have been achieved, still some key challenges and directions in this area deserve in-depth research in the future. In future research, we will focus on improving structural analysis methods for geographic networks, optimising the efficiency and accuracy of large-scale data processing tasks, expanding the construction of geographic network datasets, and deeply exploring the potential for using cau-\\sal learning in applications in this area, while also exploring cutting-edge domains and methods to further explore the potential of causal learning in the geographic domain.

\label{section7}

\sloppy
\section*{Acknowledgement}
We would like to thank the anonymous reviewers for their helpful suggestions. This research was funded by the National Natural Science Foundation of China under Grant 42271481, 42171458  and the Natural Science Foundation of Hunan Province, China under Grant 2022JJ30698. This work was carried out in part using computing resources at the High Performance Computing Platform of Central South University.



\bibliographystyle{elsarticle-num} 
\bibliography{reference}

\begin{thebibliography}{10}
\expandafter\ifx\csname url\endcsname\relax
  \def\url#1{\texttt{#1}}\fi
\expandafter\ifx\csname urlprefix\endcsname\relax\def\urlprefix{URL }\fi
\expandafter\ifx\csname href\endcsname\relax
  \def\href#1#2{#2} \def\path#1{#1}\fi

\bibitem{1}
K.~M. Curtin, Network analysis in geographic information science: Review, assessment, and projections, Cartography and geographic information science 34~(2) (2007) 103--111.

\bibitem{2}
S.~Porta, P.~Crucitti, V.~Latora, The network analysis of urban streets: a primal approach, Environment and Planning B: planning and design 33~(5) (2006) 705--725.

\bibitem{3}
M.~Barth{\'e}lemy, Spatial networks, Physics reports 499~(1-3) (2011) 1--101.

\bibitem{4}
G.~Boeing, Osmnx: New methods for acquiring, constructing, analyzing, and visualizing complex street networks, Computers, environment and urban systems 65 (2017) 126--139.

\bibitem{5}
R.~Fu, Z.~Zhang, L.~Li, Using lstm and gru neural network methods for traffic flow prediction, in: 2016 31st Youth academic annual conference of Chinese association of automation (YAC), IEEE, 2016, pp. 324--328.

\bibitem{6}
F.~Schlosser, B.~F. Maier, O.~Jack, D.~Hinrichs, A.~Zachariae, D.~Brockmann, Covid-19 lockdown induces disease-mitigating structural changes in mobility networks, Proceedings of the National Academy of Sciences 117~(52) (2020) 32883--32890.

\bibitem{7}
D.~Zhu, F.~Zhang, S.~Wang, Y.~Wang, X.~Cheng, Z.~Huang, Y.~Liu, Understanding place characteristics in geographic contexts through graph convolutional neural networks, Annals of the American Association of Geographers 110~(2) (2020) 408--420.

\bibitem{8}
S.~Hu, S.~Gao, L.~Wu, Y.~Xu, Z.~Zhang, H.~Cui, X.~Gong, Urban function classification at road segment level using taxi trajectory data: A graph convolutional neural network approach, Computers, Environment and Urban Systems 87 (2021) 101619.

\bibitem{9}
X.~Yu, S.~Shi, L.~Xu, A spatial--temporal graph attention network approach for air temperature forecasting, Applied Soft Computing 113 (2021) 107888.

\bibitem{10}
Z.~Wu, S.~Pan, F.~Chen, G.~Long, C.~Zhang, S.~Y. Philip, A comprehensive survey on graph neural networks, IEEE transactions on neural networks and learning systems 32~(1) (2020) 4--24.

\bibitem{11}
J.~Zhou, G.~Cui, S.~Hu, Z.~Zhang, C.~Yang, Z.~Liu, L.~Wang, C.~Li, M.~Sun, Graph neural networks: A review of methods and applications, AI open 1 (2020) 57--81.

\bibitem{12}
Z.~Wu, S.~Pan, F.~Chen, G.~Long, C.~Zhang, S.~Y. Philip, A comprehensive survey on graph neural networks, IEEE transactions on neural networks and learning systems 32~(1) (2020) 4--24.

\bibitem{13}
F.~Scarselli, M.~Gori, A.~C. Tsoi, M.~Hagenbuchner, G.~Monfardini, The graph neural network model, IEEE transactions on neural networks 20~(1) (2008) 61--80.

\bibitem{14}
K.~Xu, W.~Hu, J.~Leskovec, S.~Jegelka, How powerful are graph neural networks?, arXiv preprint arXiv:1810.00826 (2018).

\bibitem{15}
C.~Fan, Y.~Yang, A.~Mostafavi, Neural embeddings of urban big data reveal emergent structures in cities, arXiv preprint arXiv:2110.12371 (2021).

\bibitem{16}
A.~G. Yeh, Urban planning and gis, Geographical information systems 2~(877-888) (1999) 1.

\bibitem{17}
N.~G. Polson, V.~O. Sokolov, Deep learning for short-term traffic flow prediction, Transportation Research Part C: Emerging Technologies 79 (2017) 1--17.

\bibitem{18}
W.~E. Marshall, N.~W. Garrick, Street network types and road safety: A study of 24 california cities, Urban Design International 15 (2010) 133--147.

\bibitem{19}
Y.~Lv, Y.~Duan, W.~Kang, Z.~Li, F.-Y. Wang, Traffic flow prediction with big data: A deep learning approach, Ieee transactions on intelligent transportation systems 16~(2) (2014) 865--873.

\bibitem{20}
M.~F. Goodchild, The fundamental laws of giscience, Invited talk at University Consortium for Geographic Information Science, University of California, Santa Barbara (2003).

\bibitem{21}
L.~Anselin, What is special about spatial data?: alternative perspectives on spatial data analysis, Technical paper/National Center for Geographic Information and Analysis (89-4) (1989).

\bibitem{22}
M.~F. Goodchild, The validity and usefulness of laws in geographic information science and geography, Annals of the Association of American Geographers 94~(2) (2004) 300--303.

\bibitem{23}
Z.~Zhang, X.~Wang, Z.~Zhang, H.~Li, Z.~Qin, W.~Zhu, Dynamic graph neural networks under spatio-temporal distribution shift, Advances in neural information processing systems 35 (2022) 6074--6089.

\bibitem{24}
R.~A. Berk, An introduction to sample selection bias in sociological data, American sociological review (1983) 386--398.

\bibitem{25}
Y.~Du, J.~Wang, W.~Feng, S.~Pan, T.~Qin, R.~Xu, C.~Wang, Adarnn: Adaptive learning and forecasting of time series, in: Proceedings of the 30th ACM international conference on information \& knowledge management, 2021, pp. 402--411.

\bibitem{oodbench}
N.~Ye, K.~Li, H.~Bai, R.~Yu, L.~Hong, F.~Zhou, Z.~Li, J.~Zhu, Ood-bench: Quantifying and understanding two dimensions of out-of-distribution generalization, in: Proceedings of the IEEE/CVF Conference on Computer Vision and Pattern Recognition, 2022, pp. 7947--7958.

\bibitem{26}
E.~Dai, T.~Zhao, H.~Zhu, J.~Xu, Z.~Guo, H.~Liu, J.~Tang, S.~Wang, A comprehensive survey on trustworthy graph neural networks: Privacy, robustness, fairness, and explainability, arXiv preprint arXiv:2204.08570 (2022).

\bibitem{27}
Y.~Bengio, T.~Deleu, N.~Rahaman, R.~Ke, S.~Lachapelle, O.~Bilaniuk, A.~Goyal, C.~Pal, A meta-transfer objective for learning to disentangle causal mechanisms, arXiv preprint arXiv:1901.10912 (2019).

\bibitem{28}
J.~Liu, Z.~Shen, Y.~He, X.~Zhang, R.~Xu, H.~Yu, P.~Cui, Towards out-of-distribution generalization: A survey, arXiv preprint arXiv:2108.13624 (2021).

\bibitem{29}
J.~Wang, C.~Lan, C.~Liu, Y.~Ouyang, T.~Qin, W.~Lu, Y.~Chen, W.~Zeng, P.~Yu, Generalizing to unseen domains: A survey on domain generalization, IEEE Transactions on Knowledge and Data Engineering (2022).

\bibitem{30}
K.~Zhou, Z.~Liu, Y.~Qiao, T.~Xiang, C.~C. Loy, Domain generalization: A survey, IEEE Transactions on Pattern Analysis and Machine Intelligence 45~(4) (2022) 4396--4415.

\bibitem{31graphood}
H.~Li, X.~Wang, Z.~Zhang, W.~Zhu, Out-of-distribution generalization on graphs: A survey, arXiv preprint arXiv:2202.07987 (2022).

\bibitem{32}
Y.~Wang, K.~Yu, G.~Xiang, F.~Cao, J.~Liang, Discovering causally invariant features for out-of-distribution generalization, Pattern Recognition (2024) 110338.

\bibitem{ligr}
H.~Li, Z.~Zhang, X.~Wang, W.~Zhu, Learning invariant graph representations for out-of-distribution generalization, Advances in Neural Information Processing Systems 35 (2022) 11828--11841.

\bibitem{inrl}
H.~Li, Z.~Zhang, X.~Wang, W.~Zhu, Invariant node representation learning under distribution shifts with multiple latent environments, ACM Transactions on Information Systems 42~(1) (2023) 1--30.

\bibitem{33}
J.~Pearl, Causality, Cambridge university press, 2009.

\bibitem{34}
R.~Guo, L.~Cheng, J.~Li, P.~R. Hahn, H.~Liu, A survey of learning causality with data: Problems and methods, ACM Computing Surveys (CSUR) 53~(4) (2020) 1--37.

\bibitem{35}
B.~Sch{\"o}lkopf, J.~von K{\"u}gelgen, From statistical to causal learning, in: Proceedings of the International Congress of Mathematicians, 2022, p.~1.

\bibitem{36good}
S.~Gui, X.~Li, L.~Wang, S.~Ji, Good: A graph out-of-distribution benchmark, Advances in Neural Information Processing Systems 35 (2022) 2059--2073.

\bibitem{37}
D.~Xia, X.~Wang, N.~Liu, C.~Shi, Learning invariant representations of graph neural networks via cluster generalization, Advances in Neural Information Processing Systems 36 (2024).

\bibitem{38}
Y.~Gao, X.~Wang, X.~He, Z.~Liu, H.~Feng, Y.~Zhang, Alleviating structural distribution shift in graph anomaly detection, in: Proceedings of the Sixteenth ACM International Conference on Web Search and Data Mining, 2023, pp. 357--365.

\bibitem{39}
K.~Xu, W.~Hu, J.~Leskovec, S.~Jegelka, How powerful are graph neural networks?, arXiv preprint arXiv:1810.00826 (2018).

\bibitem{40gcn}
T.~N. Kipf, M.~Welling, Semi-supervised classification with graph convolutional networks, arXiv preprint arXiv:1609.02907 (2016).

\bibitem{41}
F.~Wu, A.~Souza, T.~Zhang, C.~Fifty, T.~Yu, K.~Weinberger, Simplifying graph convolutional networks, in: International conference on machine learning, PMLR, 2019, pp. 6861--6871.

\bibitem{42}
J.~Chen, T.~Ma, C.~Xiao, Fastgcn: fast learning with graph convolutional networks via importance sampling, arXiv preprint arXiv:1801.10247 (2018).

\bibitem{43gat}
P.~Veli{\v{c}}kovi{\'c}, G.~Cucurull, A.~Casanova, A.~Romero, P.~Lio, Y.~Bengio, Graph attention networks, arXiv preprint arXiv:1710.10903 (2017).

\bibitem{44graphsage}
W.~Hamilton, Z.~Ying, J.~Leskovec, Inductive representation learning on large graphs, Advances in neural information processing systems 30 (2017).

\bibitem{otgnn}
Q.~Zhu, Y.~Jiao, H.~Wang, N.~Ponomareva, B.~Perozzi, Gnn domain adaptation using optimal transport (2023).

\bibitem{45}
L.~Wu, H.~Lin, Y.~Huang, S.~Z. Li, Knowledge distillation improves graph structure augmentation for graph neural networks, Advances in Neural Information Processing Systems 35 (2022) 11815--11827.

\bibitem{46}
T.~Zhao, Y.~Liu, L.~Neves, O.~Woodford, M.~Jiang, N.~Shah, Data augmentation for graph neural networks, in: Proceedings of the aaai conference on artificial intelligence, Vol.~35, 2021, pp. 11015--11023.

\bibitem{47}
S.~Liu, R.~Ying, H.~Dong, L.~Li, T.~Xu, Y.~Rong, P.~Zhao, J.~Huang, D.~Wu, Local augmentation for graph neural networks, in: International conference on machine learning, PMLR, 2022, pp. 14054--14072.

\bibitem{48}
Y.~You, T.~Chen, Y.~Sui, T.~Chen, Z.~Wang, Y.~Shen, Graph contrastive learning with augmentations, Advances in neural information processing systems 33 (2020) 5812--5823.

\bibitem{49}
K.~Kong, G.~Li, M.~Ding, Z.~Wu, C.~Zhu, B.~Ghanem, G.~Taylor, T.~Goldstein, Robust optimization as data augmentation for large-scale graphs, in: Proceedings of the IEEE/CVF Conference on Computer Vision and Pattern Recognition, 2022, pp. 60--69.

\bibitem{50}
H.~Zhang, M.~Cisse, Y.~N. Dauphin, D.~Lopez-Paz, mixup: Beyond empirical risk minimization, arXiv preprint arXiv:1710.09412 (2017).

\bibitem{51}
Y.~Liu, X.~Wang, S.~Wu, Z.~Xiao, Independence promoted graph disentangled networks, in: Proceedings of the AAAI Conference on Artificial Intelligence, Vol.~34, 2020, pp. 4916--4923.

\bibitem{52}
W.~Lin, H.~Lan, B.~Li, Generative causal explanations for graph neural networks, in: International Conference on Machine Learning, PMLR, 2021, pp. 6666--6679.

\bibitem{53}
X.~Guo, L.~Zhao, Z.~Qin, L.~Wu, A.~Shehu, Y.~Ye, Interpretable deep graph generation with node-edge co-disentanglement, in: Proceedings of the 26th ACM SIGKDD international conference on knowledge discovery \& data mining, 2020, pp. 1697--1707.

\bibitem{54}
D.~Buffelli, P.~Li{\`o}, F.~Vandin, Sizeshiftreg: a regularization method for improving size-generalization in graph neural networks, Advances in Neural Information Processing Systems 35 (2022) 31871--31885.

\bibitem{55}
H.~Xue, K.~Zhou, T.~Chen, K.~Guo, X.~Hu, Y.~Chang, X.~Wang, Cap: Co-adversarial perturbation on weights and features for improving generalization of graph neural networks, arXiv preprint arXiv:2110.14855 (2021).

\bibitem{56}
S.~Li, X.~Wang, A.~Zhang, Y.~Wu, X.~He, T.-S. Chua, Let invariant rationale discovery inspire graph contrastive learning, in: International conference on machine learning, PMLR, 2022, pp. 13052--13065.

\bibitem{irm}
M.~Arjovsky, L.~Bottou, I.~Gulrajani, D.~Lopez-Paz, Invariant risk minimization, arXiv preprint arXiv:1907.02893 (2019).

\bibitem{srdo}
Z.~Shen, P.~Cui, T.~Zhang, K.~Kunag, Stable learning via sample reweighting, in: Proceedings of the AAAI Conference on Artificial Intelligence, Vol.~34, 2020, pp. 5692--5699.

\bibitem{ail}
N.~Ye, J.~Tang, H.~Deng, X.-Y. Zhou, Q.~Li, Z.~Li, G.-Z. Yang, Z.~Zhu, Adversarial invariant learning, in: 2021 IEEE/CVF Conference on Computer Vision and Pattern Recognition (CVPR), IEEE, 2021, pp. 12441--12449.

\bibitem{57dgnn}
S.~Fan, X.~Wang, C.~Shi, K.~Kuang, N.~Liu, B.~Wang, Debiased graph neural networks with agnostic label selection bias, IEEE transactions on neural networks and learning systems (2022).

\bibitem{58oodgnn}
H.~Li, X.~Wang, Z.~Zhang, W.~Zhu, Ood-gnn: Out-of-distribution generalized graph neural network, IEEE Transactions on Knowledge and Data Engineering (2022).

\bibitem{59cmrl}
N.~Lee, K.~Yoon, G.~S. Na, S.~Kim, C.~Park, Shift-robust molecular relational learning with causal substructure, in: Proceedings of the 29th ACM SIGKDD Conference on Knowledge Discovery and Data Mining, 2023, pp. 1200--1212.

\bibitem{60stablegnn}
S.~Fan, X.~Wang, C.~Shi, P.~Cui, B.~Wang, Generalizing graph neural networks on out-of-distribution graphs, IEEE Transactions on Pattern Analysis and Machine Intelligence (2023).

\bibitem{61dse}
Y.-X. Wu, X.~Wang, A.~Zhang, X.~Hu, F.~Feng, X.~He, T.-S. Chua, Deconfounding to explanation evaluation in graph neural networks, arXiv preprint arXiv:2201.08802 (2022).

\bibitem{62cal}
Y.~Sui, X.~Wang, J.~Wu, M.~Lin, X.~He, T.-S. Chua, Causal attention for interpretable and generalizable graph classification, in: Proceedings of the 28th ACM SIGKDD Conference on Knowledge Discovery and Data Mining, 2022, pp. 1696--1705.

\bibitem{63ciga}
Y.~Chen, Y.~Zhang, Y.~Bian, H.~Yang, M.~Kaili, B.~Xie, T.~Liu, B.~Han, J.~Cheng, Learning causally invariant representations for out-of-distribution generalization on graphs, Advances in Neural Information Processing Systems 35 (2022) 22131--22148.

\bibitem{64size}
B.~Bevilacqua, Y.~Zhou, B.~Ribeiro, Size-invariant graph representations for graph classification extrapolations, in: International Conference on Machine Learning, PMLR, 2021, pp. 837--851.

\bibitem{65gMPNNs}
Y.~Zhou, G.~Kutyniok, B.~Ribeiro, Ood link prediction generalization capabilities of message-passing gnns in larger test graphs, Advances in Neural Information Processing Systems 35 (2022) 20257--20272.

\bibitem{66}
T.~Zhao, G.~Liu, D.~Wang, W.~Yu, M.~Jiang, Learning from counterfactual links for link prediction, in: International Conference on Machine Learning, PMLR, 2022, pp. 26911--26926.

\bibitem{67}
Z.~Guo, J.~Li, T.~Xiao, Y.~Ma, S.~Wang, Towards fair graph neural networks via graph counterfactual, in: Proceedings of the 32nd ACM International Conference on Information and Knowledge Management, 2023, pp. 669--678.

\bibitem{68}
P.~Spirtes, Introduction to causal inference., Journal of Machine Learning Research 11~(5) (2010).

\bibitem{69}
J.~Pearl, M.~Glymour, N.~P. Jewell, Causal inference in statistics: A primer, John Wiley \& Sons, 2016.

\bibitem{80cgraphsage}
T.~Zhang, H.-R. Shan, M.~A. Little, Causal graphsage: A robust graph method for classification based on causal sampling, Pattern Recognition 128 (2022) 108696.

\bibitem{bootstrapping}
M.~A. Little, R.~Badawy, Causal bootstrapping, arXiv preprint arXiv:1910.09648 (2019).

\bibitem{70}
A.~Gretton, O.~Bousquet, A.~Smola, B.~Sch{\"o}lkopf, Measuring statistical dependence with hilbert-schmidt norms, in: International conference on algorithmic learning theory, Springer, 2005, pp. 63--77.

\bibitem{71}
K.~Fukumizu, A.~Gretton, X.~Sun, B.~Sch{\"o}lkopf, Kernel measures of conditional dependence, Advances in neural information processing systems 20 (2007).

\bibitem{72}
D.~B. Rubin, Estimating causal effects of treatments in randomized and nonrandomized studies., Journal of educational Psychology 66~(5) (1974) 688.

\bibitem{73}
N.~C. Chesnaye, V.~S. Stel, G.~Tripepi, F.~W. Dekker, E.~L. Fu, C.~Zoccali, K.~J. Jager, An introduction to inverse probability of treatment weighting in observational research, Clinical Kidney Journal 15~(1) (2022) 14--20.

\bibitem{74pubmed}
P.~Sen, G.~Namata, M.~Bilgic, L.~Getoor, B.~Galligher, T.~Eliassi-Rad, Collective classification in network data, AI magazine 29~(3) (2008) 93--93.

\bibitem{75citeseer}
A.~K. McCallum, K.~Nigam, J.~Rennie, K.~Seymore, Automating the construction of internet portals with machine learning, Information Retrieval 3 (2000) 127--163.

\bibitem{76}
Y.~Kang, S.~Gao, Y.~Liang, M.~Li, J.~Rao, J.~Kruse, Multiscale dynamic human mobility flow dataset in the us during the covid-19 epidemic, Scientific data 7~(1) (2020) 390.

\bibitem{77}
T.~Hu, S.~Wang, B.~She, M.~Zhang, X.~Huang, Y.~Cui, J.~Khuri, Y.~Hu, X.~Fu, X.~Wang, et~al., Human mobility data in the covid-19 pandemic: characteristics, applications, and challenges, International Journal of Digital Earth 14~(9) (2021) 1126--1147.

\bibitem{78}
Y.~Liang, J.~Zhu, W.~Ye, S.~Gao, Region2vec: Community detection on spatial networks using graph embedding with node attributes and spatial interactions, in: Proceedings of the 30th International Conference on Advances in Geographic Information Systems, 2022, pp. 1--4.

\bibitem{79planetoid}
Z.~Yang, W.~Cohen, R.~Salakhudinov, Revisiting semi-supervised learning with graph embeddings, in: International conference on machine learning, PMLR, 2016, pp. 40--48.

\bibitem{DANN}
Y.~Ganin, E.~Ustinova, H.~Ajakan, P.~Germain, H.~Larochelle, F.~Laviolette, M.~March, V.~Lempitsky, Domain-adversarial training of neural networks, Journal of machine learning research 17~(59) (2016) 1--35.

\bibitem{deepcoral}
B.~Sun, K.~Saenko, Deep coral: Correlation alignment for deep domain adaptation, in: Computer Vision--ECCV 2016 Workshops: Amsterdam, The Netherlands, October 8-10 and 15-16, 2016, Proceedings, Part III 14, Springer, 2016, pp. 443--450.

\bibitem{vlal}
L.~Zhu, W.~Yin, Y.~Yang, F.~Wu, Z.~Zeng, Q.~Gu, X.~Wang, C.~Zhou, N.~Ye, Vision-language alignment learning under affinity and divergence principles for few-shot out-of-distribution generalization, International Journal of Computer Vision (2024) 1--33.

\bibitem{ot}
N.~Courty, R.~Flamary, D.~Tuia, A.~Rakotomamonjy, Optimal transport for domain adaptation, IEEE transactions on pattern analysis and machine intelligence 39~(9) (2016) 1853--1865.

\end{thebibliography}





\end{document}